\documentclass[conference]{IEEEtran}

  	\usepackage[pdftex]{graphicx}
  	\graphicspath{{../pdf/}{../jpeg/}}
	\DeclareGraphicsExtensions{.pdf,.jpeg,.png}

	\usepackage[cmex10]{amsmath}
	\usepackage{algorithmic}
	\usepackage{array}
	\usepackage{mdwmath}
	\usepackage{mdwtab}
	\usepackage{eqparbox}
	\usepackage{url}
        \usepackage{lipsum}  
        \usepackage{booktabs}
        \usepackage{bm}
        \usepackage{amsmath}
        \usepackage{amssymb}
        \usepackage{algorithm}
        \usepackage{subcaption}
        \usepackage{caption}
        \usepackage{multirow}
        \usepackage{graphicx}
        \usepackage{float} 
        \usepackage{placeins} 

\begin{document}

\title{\LARGE Alignment For Performance Improvement in Conversation Bots}



 \author{\authorblockN{Raghav Garg,Kapil Sharma,Shrey Singla.}
 \authorblockA{Sprinklr India}
}

\maketitle

\begin{abstract}
This paper shows that alignment methods  can achieve  superior adherence to guardrails compared to instruction fine-tuning alone in conversational agents, also known as bots, within predefined guidelines or 'guardrails'. It examines traditional training approaches such as instruction fine-tuning and  the recent advancements in direct alignment methods like Identity Preference Optimization (IPO), and Kahneman-Tversky Optimization (KTO). The effectiveness of alignment techniques both pre and post-instruction tuning is highlighted, illustrating their potential to optimize conversational bots in domains that require strict adherence to specified rules, such as customer care.  
\end{abstract}

\IEEEoverridecommandlockouts

\IEEEpeerreviewmaketitle


\section{Introduction}
This paper delves into the burgeoning market for conversational agents known as bots, specifically those that are designed to operate within specific parameters or 'guardrails'. It explores the varying types of conversation bots, such as persona bots and dialogue tree bots, that adhere to nuanced guidelines and restrictions.Traditional training protocols for training these bots have largely relied on instruction fine-tuning based on any given dataset.

\vspace{0.3cm}
However, the recent introduction of direct alignment methods  have revolutionized the bot alignment process. This has been particularly impactful in domains where negative samples are readily available. These alignment methodologies function as efficiently as - if not slightly better than - instruction tuning alone. They are especially useful as they remove the need to train a reward model and undergo Reinforcement Learning from Human Feedback(RLHF) which was an expensive technique and also difficult to execute\cite{rafailov2023direct} and several other works followed building upon it\cite{azar2023general} \cite{ethayarajh2023halos}\cite{zhao2023slichf}.

\section{Background}
The adherence to guardrails is driven by a variety of factors such as ensuring the functionality of the bot, maintaining a specific brand voice, following ethical guidelines, upholding privacy rules, and ensuring customer satisfaction. For example, a company might establish guardrails that define that their bot should only provide responses related to their products or services and should refrain from venturing into areas that are controversial or irrelevant. 

\vspace{0.3cm}
There are different types of conversational bots that follow such guardrails. Persona bots, for instance, are developed to simulate a specific character or role. They align with their defined persona's viewpoints, communicate in a style consistent with that character, and certainly do not exceed the bounds of their persona's knowledge or attitudes. 

\vspace{0.3cm}
Dialogue Tree Bots also operate within predetermined guardrails. These bots follow a pre-constructed conversational path, making them ideal for situations where the conversation is largely predictable. This could include customer service scenarios, where there are standard responses to common requests or complaints. 

This work primarily focuses on Customer Care use cases where the companies usually have a set of rules/instructions for their Agents to follow. Currently Open-Source Commercial Models like Llama\cite{touvron2023llama},Mistral\cite{jiang2023mistral} etc. fail to adhere to there guardrails strictly and often not comply as the instruction becomes complex.

Commercial usage of these bots is typically carried forward by instruction-fine tuning on a selected dataset for the desired use-case simply because "it solves the purpose!". 
\vspace{0.3cm}

\section{Related Work}
 This ‘instruction-tuning’ procedure enables LLMs to generalize
to instructions outside of the instruction-tuning set and generally increase their usability.
However there is often accompanied by catastrophic forgetting\cite{luo2023empirical},Overfitting(loss of generalization), possible hallucination induction, model safety compromise.

\vspace{0.3cm}

Alignment Training has shown to improve helpfulness and reduce harmlessness of models.These methods first optimize a neural network reward function for compatibility with the dataset of preferences under a preference model such as the Bradley-Terry model\cite{bradley1952rank}  then fine-tune a language model to maximize the given reward using reinforcement learning algorithms, proximal policy optimization. There has been other recent alignment techniques which have shown to optimize the same loss but without a need of explicit training of a  reward model like Direct Preference Optimization(DPO)\cite{rafailov2023direct} , Identity Preference Optimization(IPO)\cite{azar2023general},Kahneman-Tversky Optimization(KTO)\cite{ethayarajh2023halos}.

\vspace{0.3cm}

We have used Alignment to improve Instruction Adherence in Conversation Bot setting where the Bot is expected to adhere follow certain Gaurdrails/playbook. There has been other works that show that Alignment(especially RLHF) \cite{ramamurthy2023reinforcement} \cite{ouyang2022training} can help to improve instruction following but ours is the first work that shows alignment can be used as an alternative to SFT in certain domains at least(where the notion of "negative" is clear) and also help in Feedback Driven Improvement all that without a need of explicitly training a reward function.

\section{Preliminaries}

In This section,we introduce various techniques that are used for alignment of LLMs.
Alignment Training aims to subtly tune the model to favour certain outputs(chosen outputs) over others(rejected responses).This is very similar to the Contrastive Learning for Embeddings where similar examples are brought closer in embedding space than dissimilar ones. 
\vspace{0.5cm}

\subsection{RLHF with PPO}
It used  Reinforcement Learning Process on PPO Loss(RLHF)\cite{schulman2017proximal}. RL was needed as PPO loss was inherently non-differentiable. The training using this technique is very expensive and highly difficult to train(unstable).

The SFT model is prompted with prompts $x$ to produce pairs of answers $(y_{1}, y_{2}) \sim \pi_{\text{SFT}}(y | x)$. These are then presented to human labelers who express preferences for one answer, denoted as $y_{w} \succ y_{l} | x$ where $y_{w}$ and $y_{l}$ denote the preferred and less preferred completion amongst $(y_1, y_2)$ respectively. The preferences are assumed to be generated by some latent reward model $r^{*}(y, x)$.

During the RL phase, we use the learned reward function to provide feedback to the language model. In particular, we formulate the following optimization problem 
\begin{equation}
\max_{\pi_{\theta}} E_{x \sim D,y \sim \pi_{\theta}(y|x)} r_{\phi}(x, y) - \beta D_{KL} \left(\pi_{\theta}(y | x) || \pi_{\text{ref}}(y | x) \right) \tag{1}
\end{equation}

\vspace{0.3cm}
where $\beta$ is a parameter controlling the deviation from the base reference policy $\pi_{\text{ref}}$, namely the initial SFT model $\pi_{\text{SFT}}$. In practice, the language model policy $\pi_{\theta}$ is also initialized to $\pi_{\text{SFT}}$. The added constraint is important, as it prevents the model from deviating too far from the distribution on which the reward model is accurate, as well as maintaining the generation diversity and preventing mode-collapse to single high-reward answers. Due to the discrete nature of language generation, this objective is not differentiable and is typically optimized with reinforcement learning. The standard approach  has been to construct the reward function $r(x, y) = r_{\phi}(x, y) - \beta(\log \pi_{\theta}(y | x) - \log \pi_{\text{ref}}(y | x))$, and maximize using PPO.

\subsection{DPO}
A major disadvantage of directly optimising the RLHF objective was that it required training a Reward Model first and then optimize the policy based on it. This was an extremely expensive process as it required manual human tagging of model outputs and training was also highly unstable and required careful tuning of optimization parameters.
The authors eliminated the need to train a reward model and show that optimising the RLHF objective is same as optimising the below objective:

\begin{equation}
\begin{split}
\textit{LDPO}(\pi_{\theta}; \pi_{ref}) = -E_{(x,y_{w},y_{l})\sim D} \Bigg[ \log \sigma \Bigg( \beta \log \frac{\pi_{\theta}(y_{w} | x)}{\pi_{ref}(y_{w} | x)} \\
 - \beta \log \frac{\pi_{\theta}(y_{l} | x)}{\pi_{ref}(y_{l} | x)} \Bigg) \Bigg].
\end{split} 
\tag{2}
\end{equation}

The general DPO pipeline is as follows: 
\begin{enumerate}
\item Sample completions $y_{1}$, $y_{2} \sim \pi_{\text{ref}}(\cdot | x)$ for every prompt $x$, label with human preferences to construct the offline dataset of preferences $D = \{(x^{(i)}, y_{w}^{(i)}, y_{l}^{(i)})\}_{i=1}^{N}$ 

\item Optimize the language model $\pi_{\theta}$ to minimize LDPO for the given $\pi_{\text{ref}}$ and $D$ and desired $\beta$. 
\end{enumerate}

In practice, one would like to reuse preference datasets publicly available, rather than generating samples and gathering human preferences. Since the preference datasets are sampled using $\pi_{\text{SFT}}$, we initialize $\pi_{\text{ref}} = \pi_{\text{SFT}}$ whenever available. However, when $\pi_{\text{SFT}}$ is not available, we initialize $\pi_{\text{ref}}$ by maximizing likelihood of preferred completions $(x, y_{w})$, that is, $\pi_{\text{ref}} = \arg \max_{\pi} \mathbb{E}_{x,y_{w} \sim D} [\log \pi(y_{w} | x)]$. This procedure helps mitigate the distribution shift between the true reference distribution which is unavailable, and $\pi_{\text{ref}}$ used by DPO.

\subsection{identity-PO}

\vspace{0.1cm}

Consider a general non-decreasing function $\Psi : [0, 1] \rightarrow \mathbb{R}$, a reference policy $\pi_{\text{ref}} \in \Delta_{X_{Y}}$, and a real positive regularisation parameter $\tau \in \mathbb{R^{*}_{+}}$, and define the $\Psi$-preference optimisation objective ($\Psi$PO) as
\begin{equation}
\max_{\pi} E_{x\sim\rho, y\sim\pi(.|x), y' \sim\mu(.|x)} [\Psi(p^{*}(y \succ y' |x))] - \tau D_{KL}(\pi || \pi_{\text{ref}}). \tag{3}
\end{equation}
This objective balances the maximisation of a potentially non-linear function of preference probabilities with the KL regularisation term which encourages policies to be close to the reference $\pi_{\text{ref}}$.

\vspace{0.3cm}

The authors here show that optimising for (3) is same as optimising for (1) under certain value of 
$\Psi$. 

They have observed in the previous section that DPO is prone to overfitting, and this stems from a combination of the unboundedness of $\Psi$, together with not training explicit reward function.A particularly natural form of objective to consider is given by taking $\Psi$ to be the
identity mapping.
\begin{algorithm}
    \caption{Sampled IPO Algorithm}
    \label{alg:sampled_ipo}
    \begin{algorithmic}[1]
        \STATE Consider a dataset $D$ of prompts, preferred and dispreferred generations $x$, $y_w$ and $y_l$, respectively. A reference policy $\pi_{\text{ref}}$.
          
        \STATE Define $h_{\pi}(y,y',x) = \log \left(\frac{\pi(y|x)\pi_{\text{ref}}(y'|x)}{\pi(y'|x)\pi_{\text{ref}}(y|x)}\right)$
          
        \STATE Starting from $\pi=\pi_{\text{ref}}$, minimize $\mathbb{E}_{(y_w,y_l,x)\sim D} \left[\left(h_{\pi}(y_w,y_l,x)-\frac{\tau^{-1}}{2}\right)^2\right]$
    \end{algorithmic}
\end{algorithm}

\vspace{0.3cm}

\begin{figure*}[!htb]
  \centering

  \begin{subfigure}[b]{0.9\textwidth}
    \includegraphics[width=\textwidth]{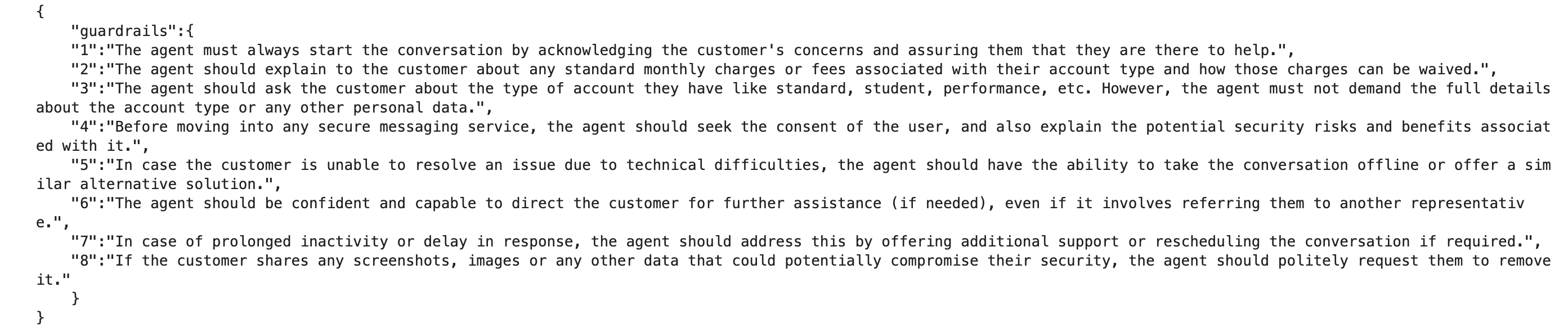}
    \caption{Guardrails sample}
    \label{fig:image1}
  \end{subfigure}
  \vspace{1em}

  \begin{subfigure}[b]{0.9\textwidth}
    \includegraphics[width=\textwidth]{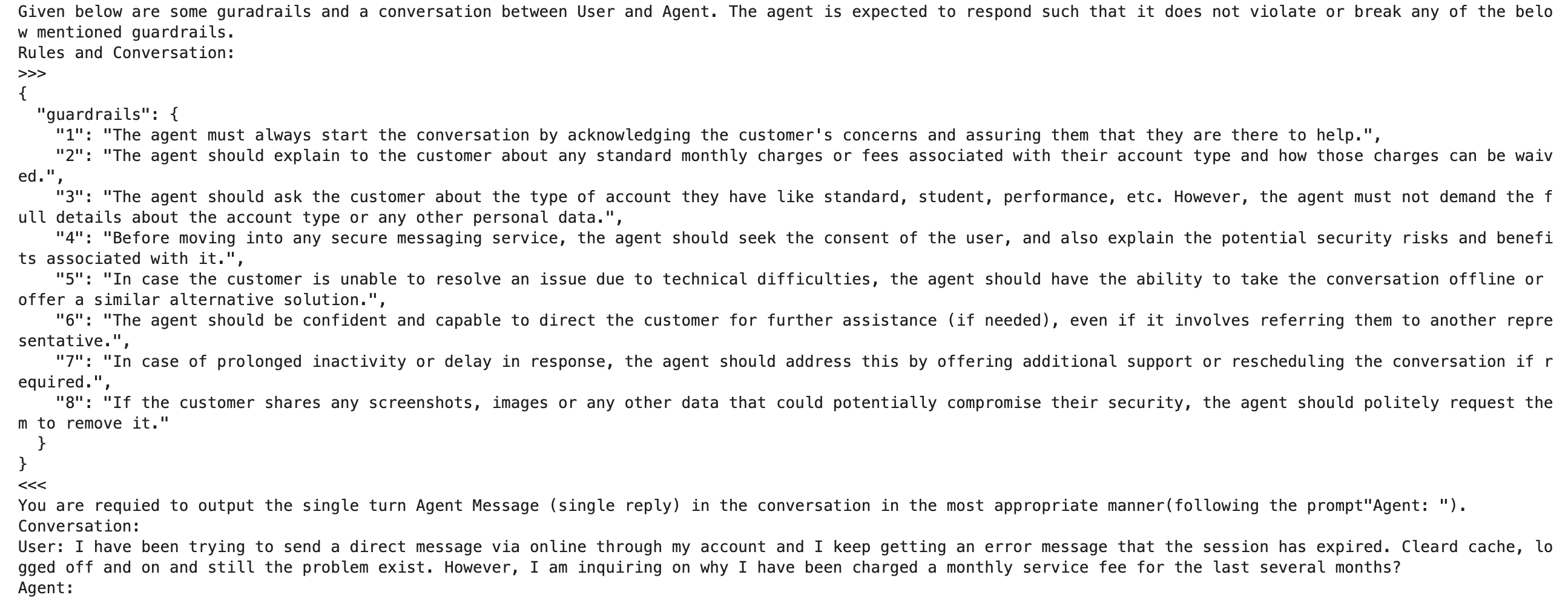}
    \caption{prompt sample}
    \label{fig:image2}
  \end{subfigure}

  \begin{subfigure}[b]{0.9\textwidth}
    \includegraphics[width=\textwidth]{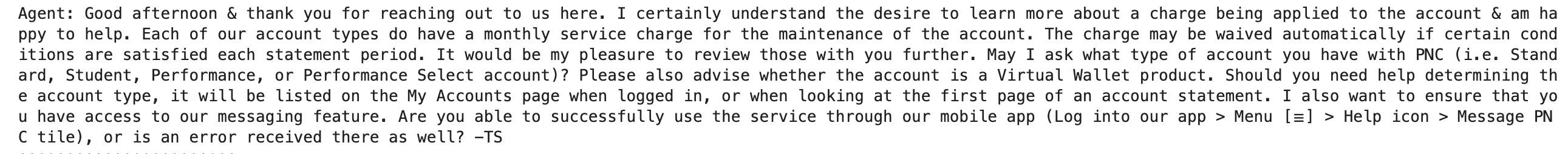}
    \caption{Chosen response}
    \label{fig:image3}
  \end{subfigure}
  \hfill

  \begin{subfigure}[b]{0.9\textwidth}
    \includegraphics[width=\textwidth]{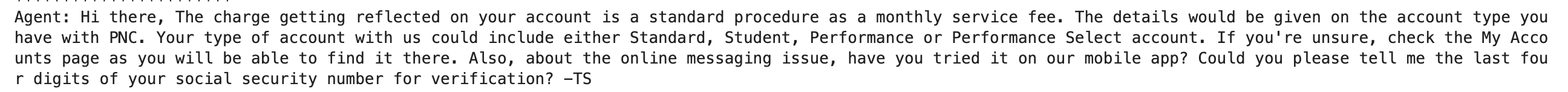}
    \caption{Rejected Response}
    \label{fig:image4}
  \end{subfigure}

  \caption{(a) shows a sample of the guardrails obtained after Stage 1 of data annotation process. Further after Stage 2 and sub sampling at agent turns, we also obtain prompts as in figure (b),a chosen response as in figure (c) and a rejected response as shown in figure (d)}
  \label{fig:images}
\end{figure*}

\subsection{kahneman-Tversky Optimization}

\vspace{0.1cm}

KTO introduces a loss function that decouples preferred and rejected outputs from  the same prompt by using HALOs(Human Centered Loss Functions).

\vspace{0.3cm}
\begin{equation}
\begin{split}
h(x, y; \beta) &= \sigma(r^*(x, y) - E_{x' \sim D, y' \sim \pi^*} [r^*(x', y')]) \\
&= \sigma\left( \beta\log\frac{\pi^*(y|x)}{\pi_{\text{ref}}(y|x)} - E_{x' \sim D} [\beta KL(\pi^* ||\pi_{\text{ref}})] \right)
\end{split}
\end{equation}

The second half of the loss function is calculated among a batch of prompts and it only needs the completion and the information whether that response is preferred or not preferred. The second half expected value is calculated by averaging the rejected batch examples for a preferred data point and averages the preferred batch samples for a rejected data point.

\section{Experiments}

\subsection{Datasets}
Although there are many datasets available that collate human preferences for training\cite{pmlr-v162-ethayarajh22a}\cite{bai2022training},they are mostly designed to optimize harmfulness and helpfulness metrics of the model. So we generate our own dataset to enhance the instruction adherence capability of the model. 
The Dataset was constructed using Fake Customer Care Conversations dataset between an Agent and a User and using GPT-IV to populate responses.
\begin{figure*}[!htbp]
  \centering
  \begin{subfigure}[b]{0.45\textwidth}
    \includegraphics[width=4cm]{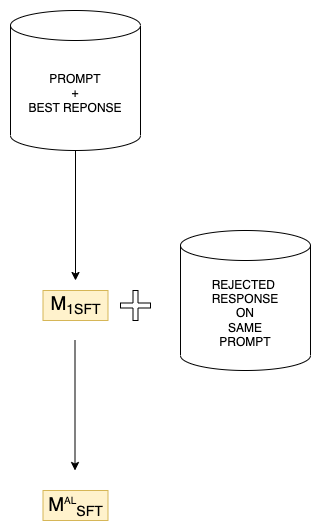} 
    \caption{The base model is instruct Fine tuned on "chosen" samples to get M1 and then it is aligned on "chosen" and "rejected" samples.}
    \label{fig:image1}
  \end{subfigure}
  \hfill
  \begin{subfigure}[b]{0.45\textwidth}
    \includegraphics[width=4cm]{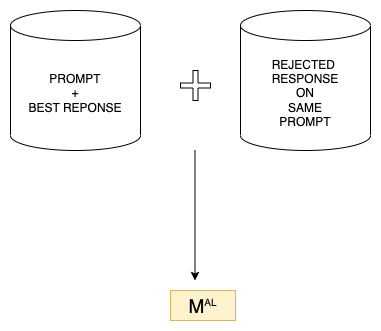} 
    \caption{The base model is directly aligned on "chosen" and "rejected" samples }
    \label{fig:image2}
  \end{subfigure}
  
  \vspace{1em}
  
  \begin{subfigure}[b]{\textwidth}
    \centering
    \includegraphics[width=4cm]{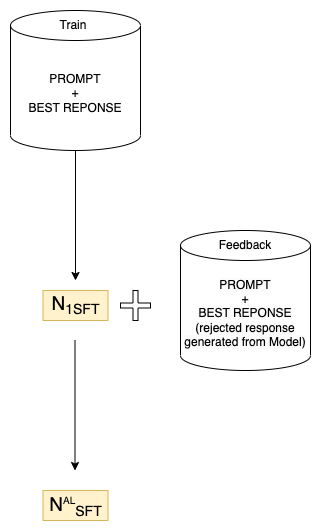} 
    \caption{The Key difference is That in the second alignment stage, the responses of the model from SFT stage are chosen as the rejected response and the feedback set is filtered to only select those examples where chosen response>models output(evaluated using GPT4 )}
    \label{fig:image3}
  \end{subfigure}
  
  \caption{Different Experiment Flows:(a) and (b) refer to Flow 1 and (c) referes to Flow 2}
  \label{fig:all_images}
\end{figure*}
\subsubsection{Stage 1}
The first stage included populating Conversation guardrails that are followed in each conversation and tagging Agent responses which follows that guardrails.
\subsubsection{Stage 2}
Then for each agent message which followed the guardrails, a negative/rejected message was constructed which clearly breaks the guardrails.

A Total of 880 conversations were tagged using the strategy above. The total number of data points increased to 8457 after breaking at each guardrail following agent message.

\vspace{0.5cm}

\subsection{Experiment Flows}
We considered Two separate experiment flows for comparing Alignment tuning vs supervised fine tuning.

Here we have considered 2 Alignment Techniques: IPO and KTO as the authors of IPO have shows it to better generalize to external dataset than DPO and we use KTO in paired mode with large batch size(using "kto\_pair" loss provided by huggingface.

\vspace{0.3cm}

\subsubsection{Flow 1}
The Dataset constructed using the above mentioned strategy was divided into a Training set and a Test Set. 

\begin{table*}[htbp]
\centering
\resizebox{0.6\textwidth}{!}{%
\begin{tabular}{|c|c|c|c|c|}
\hline
 & \multirow{2}{*}{Models} & \multicolumn{3}{c|}{Overall Win Rate} \\ \cline{3-5}
 &  & Adherence & Naturalness & Hallucination \\ \hline
\multirow{4}{*}{Flow 1} & ${M}_{{1}{SFT}}$ & $70.4\%$ & $47.7\%$ & {\boldmath$95.8\%$} \\ 
 & ${M}_{{1}{SFT}}^{{ipo}}$ & {\boldmath$76.9\%$} & {\boldmath$65.5\%$} & $92.8\%$ \\ \cmidrule{2-5}
 & ${M}_{{2}{SFT}}$ & $69.2\%$ & $50.6\%$ & \boldmath$94.7\%$ \\
 & ${M}_{{1}{SFT}}^{{ipo}}$ & \boldmath$74.2\%$ & \boldmath$64.6\%$ & $91.4\%$ \\ \hline
\multirow{2}{*}{Flow 2} & ${N}_{{1}{SFT}}$ & $72.3\%$ & $58.1\%$ & {\boldmath$98.2\%$}  \\
 & ${N}_{{1}{SFT}}^{{ipo}}$ & {\boldmath$79.4\%$}  & {\boldmath$76.4\%$}  & $93.3\%$ \\ \hline
\end{tabular}%
}
\caption*{\textbf{Table 1:} Comparing Overall Win Rates of Different Models } 
\end{table*}

We Created 3 different Model Variants  with their description as below:

\begin{itemize}
  \item Model Obtained after SFT on chosen responses
  \begin{itemize}
    \item $\mathbf{M}_{\mathbf{1}\textbf{SFT}}$: 1 epoch fine tuning
    \item $\mathbf{M}_{\mathbf{2}\textbf{SFT}}$: 2 epochs fine tuning
  \end{itemize}
  \item Model Obtained after direct alignment on chosen responses on Base Mistral Model(1 epoch only)
  \begin{itemize}
    \item $\mathbf{M}^{\textbf{ipo}}$: IPO Alignment
    \item $\mathbf{M}^{\textbf{kto}}$: KTO Alignment
  \end{itemize}
  \item Model Obtained after 1 epoch of SFT and 1 epoch of alignment
  \begin{itemize}
    \item $\mathbf{M}_{\mathbf{1}\textbf{SFT}}^{\textbf{ipo}}$: IPO Alignment
    \item $\mathbf{M}_{\mathbf{1}\textbf{SFT}}^{\textbf{kto}}$: KTO Alignment
  \end{itemize}
\end{itemize}

\vspace{0.5cm}

\subsubsection{Flow 2}
The Dataset was divided into three sections:Training set,a Feedback Set and a Test set. We Created 2 different Model Variants with their description as below:
\begin{itemize}
    \item $\mathbf{N}_{\mathbf{1}\textbf{SFT}}$: Model obtained after 1 epoch of instruction finetuning        
\end{itemize}

\vspace{0.5cm}

\begin{itemize}
    \item Model obtained after 1 epoch of alignment Training on M4 on a filtered feedback Dataset
    \begin{itemize}
        \item $\mathbf{N}_{\mathbf{1}\textbf{SFT}}^{\textbf{ipo}}$: IPO Alignment
        \item $\mathbf{N}_{\mathbf{1}\textbf{SFT}}^{\textbf{kto}}$: KTO Alignment
    \end{itemize}
\end{itemize}

\vspace{0.2cm}

\subsection{Technical Details}
Mistral-7B-Instruct model is used as the base model in both the experiment flows as it is released with a commercial license and it seemed to perform better than Llama-2 series in our internal experiments.Also,Learning rate is a very important factor for different training techniques.

Learning Rate for alignment Training has to be kept much lower than SFT Training otherwise it leads to repetitive outputs from Model. We used these learning rates:
\begin{itemize}
    \item SFT: 5e-4
    \item IPO:2e-6
    \item KTO:5e-7
\end{itemize}

Batch size of 8 was used to do Training using Sharding across 4 A100x80 Gbs.

\vspace{0.4cm}

\subsection{Results}

For evaluation of the Agent Responses, we use GPT4 and ask it to compare two model and rate the results in 4 bins: Both Results are Acceptable,None are acceptable, Model 1 is better or 2 is better. Rating was calculated across 3 different Dimensions : \textbf{Adherence},\textbf{Naturalness}  and \textbf{Hallucination} .

\vspace{0.3cm}

\textbf{Adherence} Tends to capture if the agent broke any of the guardrails.

\textbf{Naturalness} ensures the response follows the conversation context and is a coherent continuation.

\textbf{Hallucination} on the other hand tries to capture if the model used any outside information than the one provided in answering.

\vspace{0.3cm}

Results of both experiment Flows are given in Fig4 and Fig3 respectively.

\vspace{0.5cm}

\begin{figure*}[!htbp]
  \section*{Section 1} 
  \centering
  \begin{subfigure}[b]{\textwidth}
    \includegraphics[width=\textwidth]{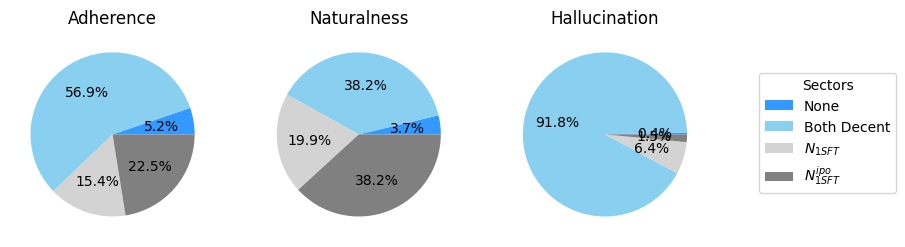}
    \caption{Model Comparison after Fine-Tuning and after Alignment Stage}
    \label{fig:image1}
  \end{subfigure}

  \section*{Section 2}
  \centering
  \begin{subfigure}[b]{\textwidth}
    \includegraphics[width=\textwidth]{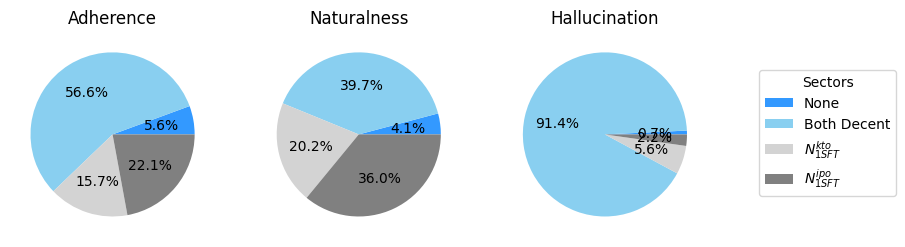}
    \caption{Model Comparison when applied KTO and IPO as alignment Techniques on the SFT model}
    \label{fig:image2}
  \end{subfigure}

  \caption{Win Rates Of Different Model in Experiment Flow 2.(a) signifies the performance gain obtained from doing alignment after the SFT stage whereas (b) shows that IPO performed better in our experiments. }
  \label{fig:all_images}
\end{figure*}

\subsubsection{\textbf{Flow 1 Results}}
Graph (a) shows that performing Alignment Training post Supervised Fine tuning has clear Benefits for Naturalness and Adherence. Similar Trends are observed for Graphs (b) and (c).

\vspace{0.1cm}

Graph (b) in Fig4, aimed to compare by replacing the second stage of alignment with another fine-tuning stage and graph (c) replaced the first stage itself with Alignment. Alignment got a win rate advantage of roughly $5\%$ in adherence and roughly $15\%$ in Naturalness.

However for Hallucination, the scores were roughly similar with both models doing good. SFT though seemed to perform a little better with an error range of $1\%$. The pattern holds for all graphs (a),(b),(c) and (d).

We also observe that IPO loss seems to work  better than KTO pair loss but however the thing to note that the true advantage of KTO is in the regime where preferred and dispreffered samples wont occur together and it might work better with large batch sizes.

\vspace{1cm}

\subsubsection{\textbf{Flow 2 Results}}
 Here the goal was to Test iterative improvement of performance of our model using a separate feedback Set as the Model outputs after SFT were kept as "rejected" responses.
 Graph (a) in Fig3 shows the performance improvement we got on our Test set by aligning our model on the Feedback set as the performance gain in Adherence is roughly $7\%$ and roughly $20\%$ in Naturalness.

 This shows that we can improve the model on additional datasets in an online fashion as usually performing SFT may leads to catastrophic forgetting but alignment usually work with much lower learning rates (so tweaking of model weights is low) and also directly minimises the KL loss with reference model which minimizes the chances of forgetting the original Training Data.
 
\vspace{0.1cm}

 Patterns in Hallucination Metrics and Graph (b) in Fig3 are same as those observed in Flow 1. Complied Results are given in the table below:

\vspace{1cm}

\section{Conclusion}
Alignment Training via IPO seemed to perform better/at par with Instruction Fine Tuning when aiming to improve Instruction Adherence in Conversation Bots. Alignment works with much lower learning rate and also optimises for distribution loss with reference model thus is good for iterative improvement for tasks like
\begin{itemize}
    \item feedback Driven improvement
    \item Safety alignment
\end{itemize}

\vspace{0.3cm}

Although other work has shown that alignment applied before/in-place of SFT does not provide same performance improvement,we believe the reasoning behind our performance gain lies in our problem setting itself. We are trying to generate guardrails driven Agent responses which have an obvious chosen answer and a rejected response( when agent breaks the guardrails). Alignment loss is more ideal for such a setting when we are trying to teach the model to NOT DO certain things and prefer some responses over others.

\vspace{0.3cm}

Also Future work can dive down into reducing Hallucination when using alignment techniques or by creating appropriate dataset or expanding domain to other tasks in customer care like generalised intent detection, insights etc. apart from conversation bots.\cite{853049}

\begin{figure*}[!htbp]
  \section*{Section 1} 
  \centering
  \begin{subfigure}[b]{\textwidth}
    \includegraphics[width=\textwidth]{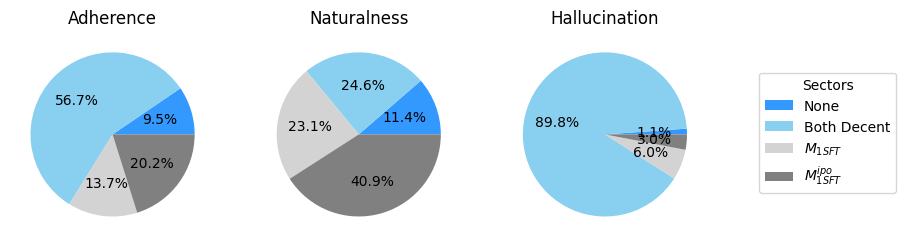}
    \caption{Model Comparison after Fine-Tuning and after Alignment Stage}
    \label{fig:image1}
  \end{subfigure}

  \section*{Section 2}
  \centering
  \begin{subfigure}[b]{\textwidth}
    \includegraphics[width=\textwidth]{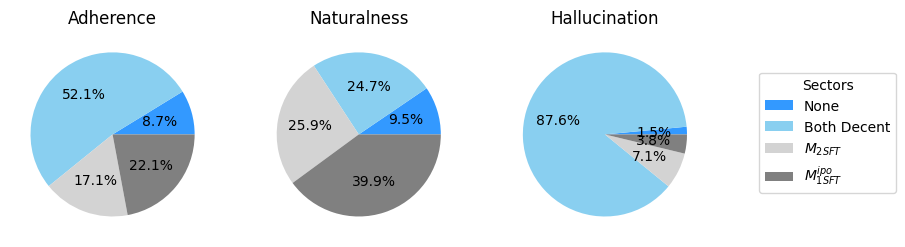}
    \caption{Model comparison if using 2 epochs of Fine-tuning vs 1 epoch of SFT and alignment}
    \label{fig:image2}
  \end{subfigure}

  \section*{Section 3}
  \centering
  \begin{subfigure}[b]{\textwidth}
    \includegraphics[width=\textwidth]{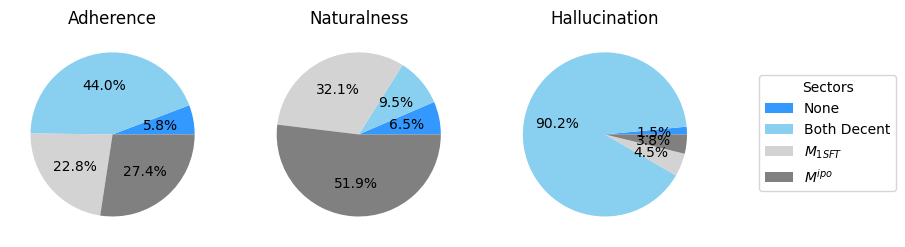}
    \caption{Model comparison if using 1 epoch of Fine-tuning vs 1 epoch of alignment}
    \label{fig:image3}
  \end{subfigure}

  \section*{Section 4}
  \centering
  \begin{subfigure}[b]{\textwidth}
    \includegraphics[width=\textwidth]{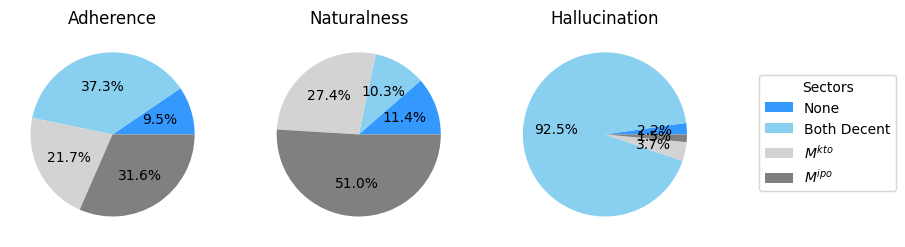}
    \caption{Model Comparison when applied KTO and IPO as alignment Techniques on the base model}
    \label{fig:image4}
  \end{subfigure}

  \caption{Win Rate Of Different Model in Experiment Flow 1. (a) and (d) show similar trends as observed in Flow 1. (b) and (c) additionally show that Alignment works superior to SFT.}
  \label{fig:all_images}
\end{figure*}

\bibliographystyle{IEEEtran}
\bibliography{IEEEabrv,main}

\end{document}